\documentclass{article}
\usepackage{ijcai21}

\usepackage{times}

\usepackage{soul}
\usepackage{url}
\usepackage[hidelinks]{hyperref}
\usepackage[utf8]{inputenc}
\usepackage[small]{caption}
\usepackage{graphicx}
\usepackage{amsmath}
\usepackage{booktabs}
\usepackage{multirow}
\usepackage{array}
\title{Person Search Challenges and Solutions: A Survey}

\author{
Xiangtan Lin$^1$\and
Pengzhen Ren$^2$\and
Yun Xiao$^2$ \and
Xiaojun Chang$^1$\footnote{Corresponding Author} \and
Alex Hauptmann$^3$

\affiliations
$^1$Monash University~~$^2$Northwest University~~$^3$Carnegie Mellon University
\emails
\{xiangtan.lin, cxj273, alex.hauptmann\}@gmail.com, pzhren@foxmail.com, yxiao@nwu.edu.cn
}

\begin{document}

\maketitle

\begin{abstract}
Person search has drawn increasing attention due to its real-world applications and research significance. Person search aims to find a probe person in a gallery of scene images with a wide range of applications, such as criminals search, multi-camera tracking, missing person search, etc. Early person search works focused on image-based person search, which uses person image as the search query. Text-based person search is another major person search category that uses free-form natural language as the search query. Person search is challenging, and corresponding solutions are diverse and complex. Therefore, systematic surveys on this topic are essential. This paper surveyed the recent works on image-based and text-based person search from the perspective of challenges and solutions. Specifically, we provide a brief analysis of highly influential person search methods considering the three significant challenges: the discriminative person features, the query-person gap, and the detection-identification inconsistency. We summarise and compare evaluation results. Finally, we discuss open issues and some promising future research directions.
\end{abstract}

\section{Introduction}
\begin{figure}[t]
\begin{center}
\includegraphics[width=\linewidth]{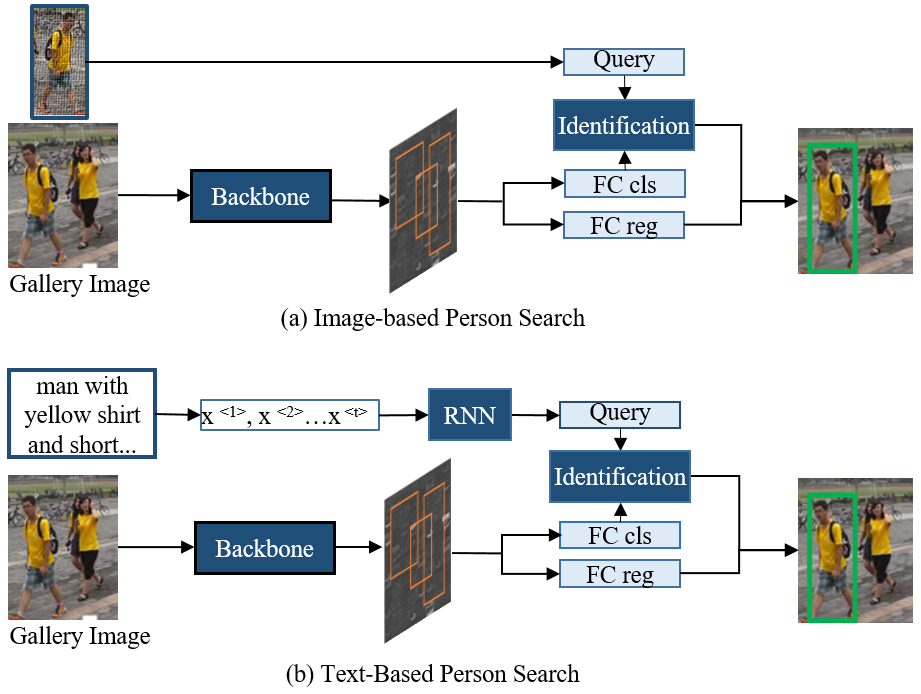}
\end{center}
\caption{The general frameworks of person search. (a) Image-based person search in which person image is available as the search query against a gallery of images. Image-based person search involves two sub-tasks, person detection and person identification. (b) Text-based person search in which search query is free form natural language. A general text-based person search framework typically learns text feature through an RNN variant network and then align text features with visual elements from the detection network to identify the person in the target images.}
\label{fig:ps}
\end{figure}

Person search~\cite{xu_person_2014} aims to find a query person in a gallery of scene images. Historically, person search was an extended form of person re-identification (re-id) problem~\cite{LiuCCPZYH20,LiuCS20,LiLCYPZ19,ChengGCSHZ18,LiuCCY18,ChengCLHGZ17,LiuC0Y17}. Therefore, early researches on person search focused on an image-based setting, which uses person image as the search query ~\cite{xiao_joint_2017,liu_neural_2017,chang_rcaa_2018,gao_structure-aware_2019,xiao_ian_2019}. Meanwhile, research in text-based person search ~\cite{li_person_2017,wang_vitaa_2020} has made significant advances in the past few years. Text-based person search is handy when a probe image is unavailable but free-form natural language. The two types of person search are illustrated in Figure \ref{fig:ps}.

\begin{figure}[t]
\begin{center}
\includegraphics[width=0.9\columnwidth]{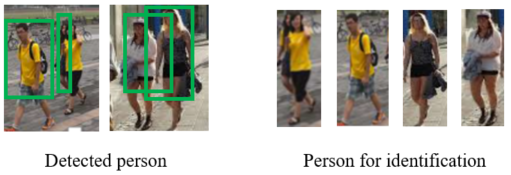}
\end{center}
\caption{The detection-identification inconsistency problem. The detecion model learns person proposal based on common person apperance using intersection-over-union (IoU) over certain threshold, which may result in less accurate bounding boxes compare to person for the identification task.}
\label{fig:inconsistency}
\end{figure}

Person search faces more challenges than person re-id problem. Unlike the person re-id setting where the cropped person images are provided, and the primary challenge is just to bring the query-person gap. Person search needs to deal with an additional detection challenge so that the detected person can be used for the downstream identification task. The additional detection task poses more challenges due to the influences of poses, occlusion, resolution and background clutter in the scene images. Such detection results may be inconsistent with the identification task (Figure \ref{fig:inconsistency}). Similarly, text-based person search is also more challenging than the traditional text-image matching problem~\cite{li_person_2017} as it needs to learn discriminative features first before the text-person matching. 

Person search is fast-evolving, and existing person search methods are diverse and complex. Researchers may leverage the rich knowledge concerning object detection, person re-id, and text-image matching separately. Systematic surveys concerning person search bring more values to the community. Especially, as far as we know, there is no existing survey covering the text-based person search. ~\cite{islam_person_2020} surveyed works on image-based person search and neglected the text-based person search. Furthermore, ~\cite{islam_person_2020} didn't discuss the joint challenge of person detection and identification, especially the detection-identification inconsistency challenge as illustrated in Figure \ref{fig:inconsistency}. Therefore, we survey works beyond image-based person search and provide a systematic review of the diverse person search solutions. We summarise the main differences between the previous survey~\cite{islam_person_2020} and ours in Table \ref{table:surveydiff}.

\begin{table} [t]
\renewcommand{\arraystretch}{1}
\begin{center}
\resizebox{\columnwidth}{!}{
\setlength{\tabcolsep}{2pt}
\begin{tabular}{l|m{0.24\columnwidth}|m{0.56\columnwidth}}
\noalign{
\global\dimen1\arrayrulewidth
\global\arrayrulewidth1pt
}\hline
\noalign{
\global\arrayrulewidth\dimen1 
}
{\bf Survey} &{\bf Covering}& {\bf Analysis}\\
\hline
~\cite{islam_person_2020} & Image-based & Components \\
\hline
Ours & Image-based, Text-based & 
(Challenges: {\bf Solutions})

Discriminative person features: {\bf Deep  feature representation learning}

Query-person gap: {\bf Deep metric learing}

Detection-identification inconsistency: {\bf Identity-driven detection}

\\
\noalign{
\global\dimen1\arrayrulewidth
\global\arrayrulewidth1pt
}\hline
\noalign{
\global\arrayrulewidth\dimen1 
}
\end{tabular}
}
\caption{Summary of the main differences between the previous survey and ours. This survey focuses more on challenges and solutions.}
\label{table:surveydiff}
\end{center}
\vspace{-4mm}
\end{table}

In this survey, we aim to provide a cohesive analysis of the recent person search works so that the rationals behind the ideas can be grasped to inspire new ideas. Specifically, We surveyed recently published and pre-print person search papers from top conference venues and journals. We analyse methods from the perspective of challenges and solutions and summarise evaluation results accordingly. At the end of the paper, we provide insights on promising future research directions.
In summary, the main contributions of this survey are:
\begin{itemize}
\item In addition to image-based person search, we cover text-based person search which was neglected in the previous person search survey.
\item We analyse person search methods from the perspective of challenges and solutions to inspire new ideas.
\item We summarise and analyse existing methods' performance and provide insights on promising future research directions.
\end{itemize}

\section{Person Search}
\begin{figure*}[t]
\begin{center}
\includegraphics[width=\textwidth]{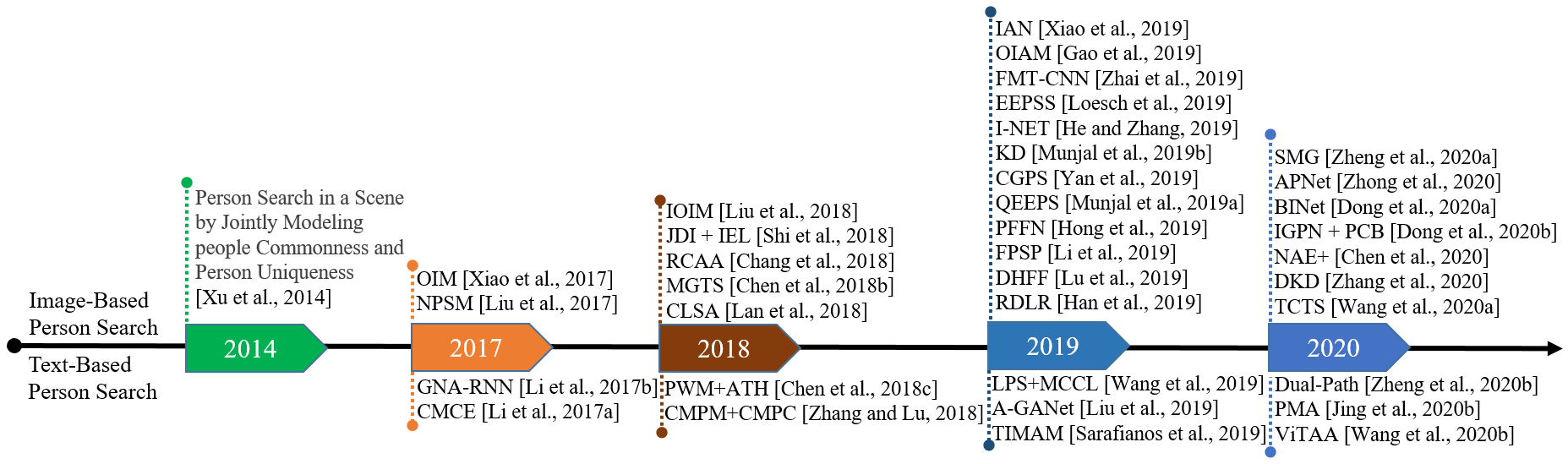}
\end{center}
\caption{Timeline of person search studies. Above the timeline are image-based person search works. Below the line are text-based person search methods.}
\label{fig:timeline}
\end{figure*}

Person search is a fast-evolving research topic. In 2014, ~\cite{xu_person_2014} first introduced the person search problem and pointed out the conflicting nature between person detection and person identification sub-tasks. Person detection deals with common human appearance, while the identification task focuses on a person's uniqueness. After ~\cite{xiao_joint_2017} introduced the first end-to-end person search framework in 2017, we have seen an increasing number of image-based person search works in the last three years. Meanwhile, in 2017, GNA-RNN~\cite{li_person_2017} set the benchmark for text-based person search. We draw a timeline to present the person search works in Figure \ref{fig:timeline} and show the two divisions: image-based and text-based person search. 

Person search addresses person detection and person identification simultaneously. There are three significant person search challenges to be considered when developing a person search solution. Firstly, a person search model needs to learn discriminative person features from scene images suitable for matching the query identity. Inevitably, the learnt person features differ from the query identity features to some degrees. Therefore the second major challenge is how to bring the gap between the query and the detected person. The third challenge is related to the conflicting nature between person detection and person identification. Person detection deals with common person appearance, while the identification task focuses on a person's uniqueness. The detected person may not be suitable for identity matching. For instance, a partial human body could be considered a person during detection and is inconsistent with the query identity at the identification stage, which may be a full person picture.

In this section, we analyse person search methods regarding above-mentioned three challenges and corresponding solutions from the following three aspects for both image-based and text-based person search:
\begin{itemize}
\item {\bf Deep feature representation learning.} Addressing the challenge of learning discriminative person features from gallery images concerning background clutter, occlusion and poses etc.
\item {\bf Deep metric learning.} Addressing the challenge of bringing query-person gap by using loss functions to guide feature representation learning.
\item {\bf Identity-driven detection.} Addressing the challenge of mitigating the detection-identification inconsistency by incorporating query identities into the detection process. 
\end{itemize}

\subsection{Deep Feature Representation Learning}
Deep feature representation learning focuses on learning discriminative person features concerning distractors in the gallery images. The majority of the early methods exploited global person features, including context cues, while refining person proposals. Such as RCAA ~\cite{chang_rcaa_2018} utilises the relational spatial and temporal context in a deep reinforcement learning framework to adjust the bounding boxes constantly. However, these methods didn't consider the background clutter in the proposal bounding boxes, resulting in a situation where different persons with similar backgrounds are close in the learnt feature space. SMG ~\cite{zheng_segmentation_2020} eliminates background clutter using segmentation masks so that the learnt person features are invariant to the background clutter. NAE~\cite{chen_norm-aware_2020} separates persons and background by norms and discriminates person identities by angles. Person detection and object detection, in general, face the multi-scale matching challenge. To learn scale-invariant features, CLSA~\cite{lan_person_2018} and DHFF ~\cite{lu_dhff_2019} utilise multi-level features from the identification network to solve the multi-scale matching problem with different multi-metric losses.

Local discriminative features are useful when two persons exhibit similar appearance and can't be discriminated against merely by full-body appearance. APNet ~\cite{zhong_robust_2020} divides the body into six parts and uses an attention mechanism to weigh the body parts' contribution further. Unlike APNet, which uses arbitrary body parts, CGPS~\cite{yan_learning_2019} proposes a region-based feature learning model for learning contextual information from a person graph. BINet ~\cite{dong_bi-directional_2020} uses the guidance from the cropped person patches to eliminate the context influence outside the bounding boxes. 

Deep feature representation learning in text-based person search learns visual representations for the detected person most correspondent to the textual features. Similar to image-based person search, text-based person search methods exploit global and local discriminative features. GNA-RNN~\cite{li_person_2017} exploits global features in the first text-based LSTM-CNN person search framework and uses an attention mechanism to learn the most relevant parts. GNA-RNN only attends to visual elements and doesn't address various text structure. To address this problem, CMCE~\cite{li_identity-aware_2017} employs a latent semantic attention module and is more robust to text syntax variations. To address the background clutter problem, PMA~\cite{jing_pose-guided_2020} uses pose information to learn the pose-related features from the map of the key points of human. To further distinguish person with similar global appearance, PWM+ATH~\cite{chen_improving_2018-1} utilises a word-image patch matching model to capture the local similarities. ViTAA~\cite{wang_vitaa_2020} decomposes both image and text into attribute components and conducts a fine-grained matching strategy to enhance the interplay between image and text.  
\subsection{Deep Metric Learning}
Deep metric learning tackles the query-person gap challenge with loss functions to guide the feature representation learning. The general purpose is to bring the detected person features close to the target identity while separating them from other identities. Similarity metrics such as Euclidean distance and cosine similarity are common measures to evaluate the similarity level among those query-person pairs. The identification task is generally formulated as a classification problem where conventional softmax loss trains the classifier. Softmax has a major problem of slow convergence with a large number of classes. OIM (Eq: \ref{eq:oimloss})~\cite{xiao_joint_2017} addresses this issue while exploiting large number of identities and unlabeled identities. OIAM~\cite{gao_structure-aware_2019} and IEL ~\cite{shi_instance_2018} further improve the OIM method with additional center losses. Different from OIM variances, I-Net~\cite{he_end--end_2019} introduces a Siamese structure with an online pairing loss (OPL) and hard example priority Softmax loss (HEP) to bring the query-person gap. RDLR~\cite{han_re-id_2019} uses the identification loss instead of regression loss for supervising the bounding boxes. 

In the landmark OIM approach, the OIM loss effectively closes the query-person gap utilising labelled and unlabeled identities from training data. The probability of detected person features $x$ being recognised as the identity with class-id $i$ by a Softmax function:  
\begin{equation} \label{eq:oimprob}
p_i = \frac{\exp(v_i^Tx/\tau)}{\sum_{j=1}^L\exp(v_j^Tx/\tau)+ \sum_{k=1}^Q\exp(u_k^Tx/\tau)}.
\end{equation}
Where $v_i^T$ is the labelled person features for the $i_{th}$ identity in the lookup table (LUT). $v_j^T$ is the $j_{th}$ labelled person features in the LUT. $u_k^T$ is the $k_{th}$ unlabelled person features in the LUT. $\tau$ regulates probability distribution. OIM objective is to maximize the expected log-likelihood of the target $t$.
\begin{equation} \label{eq:oimloss}
\mathcal{L} = \mathrm{E_x}\left[ \log{p_t}\right]. 
\end{equation}

Metric learning in text-based person search is to close the text-image modality gap. The main challenge in text-based person search is that it requires the model to deal with the complex syntax from the free-form textual description. To tackle this, methods like ViTAA, CMCE, PWM+ATH~\cite{wang_vitaa_2020,li_identity-aware_2017,chen_improving_2018-1} employ attention mechanism to build relation modules between visual and textual representations. Unlike the above three methods, which are all the CNN-RNN frameworks, Dual Path~\cite{zheng_dual-path_2020} employs CNN for textual feature learning and proposes an instance loss for image-text retrieval. CMPM+CMPC~\cite{zhang_deep_2018} utilizes a cross-modal projection matching (CMPM) loss and a cross-modal projection classification (CMPC) loss to learn discriminative image-text representations. Similar to CMPM+CMPC, MAN~\cite{jing_cross-modal_2020} proposes cross-modal objective functions for joint embedding learning to tackle the domain adaptive text-based person search.

\begin{figure}[t]
\begin{center}
\includegraphics[width=\linewidth]{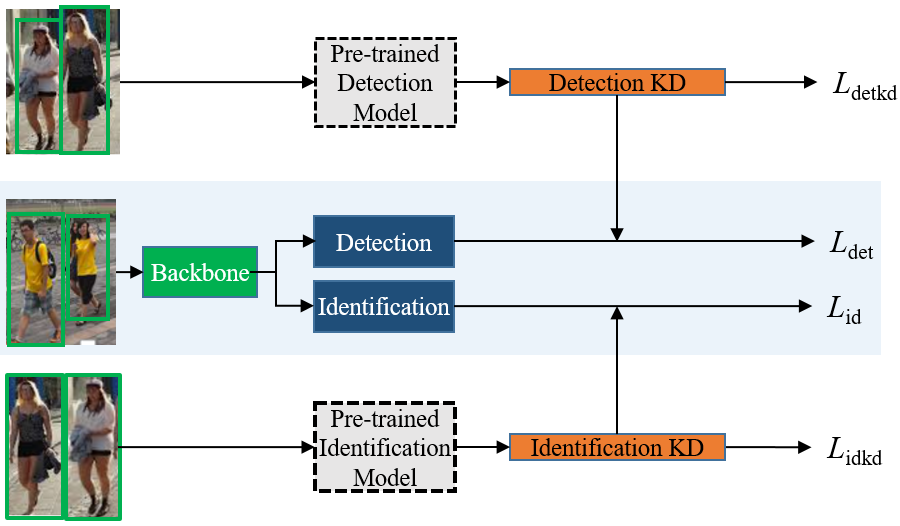}
\end{center}
\caption{A representative end-to-end person search framework where the detection and identification branches are supervised by pre-trained detection and identification models through knowledge distillation. The detection loss $L_{det}$, identification loss $L_{id}$ and the knowledge distillation losses $L_{detkd}$ and $L_{idkd}$ can be optimised as a multi-task learning task through back-propagation.}
\label{fig:kd}
\end{figure}

Inspired by the recent success of knowledge distillation~\cite{lhinton_distiling_2015}, instead of directly training detection and identification sub-nets, the two modules can be learnt from the pre-trained detection and identification models~\cite{munjal_knowledge_2019}. DKD~\cite{zhang_diverse_2020} focuses on improving the performance of identification by introducing diverse knowledge distillation in learning the identification model. Specifically, a pre-trained external identification model is used to teach the internal identification model. A simplified knowledge distillation process is illustrated in Figure \ref{fig:kd}.
 
\subsection{Identity-driven detection}\label{architecture}
The detection-identification inconsistency challenge in image-based person search is tackled by incorporating identities into the detection process. This means during training, ground-truth person identities are used to guide person proposals, or at search time, the query identity information is utilised to refine the bounding boxes. Person search tackles person detection and person identification challenges in one framework. Existing person search methods can be divided into two-stage and end-to-end solutions from the architecture perspective. In two-stage person detection, the detection and identification models are trained separately for optimal performance of both detection and identification models ~\cite{zhang_diverse_2020,loesch_end--end_2019}. However, due to the detection-identification inconsistent issue, the separately trained models may not yield the best person search result. To address the inconsistency problem between the two branches, TCTS~\cite{wang_tcts_2020} and IGPN+PCB~\cite{dong_instance_2020} exploit query information at search time to filter out low probable proposals. End-to-end methods share visual features between detection and identification and significantly decrease runtime. However, joint learning contributes to sub-optimal detection performance~\cite{wang_tcts_2020}, which subsequently worsen the detection-identification inconsistency problem. To address the problem. NPSM~\cite{liu_neural_2017} and QEEPS ~\cite{munjal_query-guided_2019} leverage query information to optimise person proposals in detection process. Differ from the query-guided methods, RDLR~\cite{han_re-id_2019} supervises bounding box generation using identification loss. Therefore, proposal bounding boxes are more reliable. In person search settings, the query identity is present in gallery images. Therefore, all methods mentioned above essentially incorporate identities into the detection process.

Text-based person search faces less detection-identification inconsistency challenge since the proposal person is identified by text-image matching without comparing bounding boxes. Therefore, text-based person search mainly focuses on learning visual and language features and improving the matching accuracy. The majority of current text-based person search methods are end-to-end frameworks that consist of a CNN backbone for extracting visual elements and a bi-LSTM for learning language representations. The two modules are jointly trained to build word-image relations from the learnt visual and language feature representations. CMCE~\cite{li_identity-aware_2017} is the only two-stage framework in which the stage-one CNN-LSTM network learns cross-modal features, and in stage-two, the CNN-LSTM network refines the matching results using an attention mechanism.
\begin{table*} [t]
\begin{center}
\resizebox{\textwidth}{!}{
\begin{tabular}{l|c|c|c|c|c|c|c|c}
\multicolumn{1}{c}{}& \multicolumn{1}{c}{}&\multicolumn{1}{c}{}&\multicolumn{2}{c}{{\bf CUHK-SYSU}} & \multicolumn{2}{c}{{\bf PRW}} & \multicolumn{2}{c}{{\bf LSPM}}\\
\noalign{
\global\dimen1\arrayrulewidth
\global\arrayrulewidth1pt
}\hline
\noalign{
\global\arrayrulewidth\dimen1 
}
{\bf Method} & {\bf Feature} & {\bf Loss} & mAP(\%) & R@1(\%) & mAP(\%) & R@1(\%)& mAP(\%) & R@1(\%)\\
\hline
\multicolumn{8}{c}{\textbf{\textit {Non-identity-driven detection}}}\\
\hline
OIM~\cite{xiao_joint_2017} & global &OIM&75.5 & 78.7 & 21.3 & 49.9 & 14.4 & 47.7 \\
IAN~\cite{xiao_ian_2019} & global & Softmax, Center loss&76.3 & 80.1& 23.0 & 61.9 &&\\
OIAM~\cite{gao_structure-aware_2019} & global &OIM, Center loss& 76.98 & 77.86 & {\bf 51.02} & 69.85 &&\\
FMT-CNN~\cite{zhai_fmt_2019} & global &OIM, Softmax& 77.2 & 79.8 & &&&\\
ELF16~\cite{yang_enhanced_2017} & global \& local &OIM& 77.8 & 80.6 &&&&\\
IOIM~\cite{liu_discriminatively_2018} &global&IOIM, Certer loss& 79.78 & 79.90 & 21.00 & 63.10 &&\\
EEPSS~\cite{loesch_end--end_2019} &global&Triplet loss& 79.4 & 80.5 & 25.2 & 47.0 &&\\
JDI + IEL  ~\cite{shi_instance_2018} & global &IEL& 79.43 & 79.66 & 24.26 & 69.47 &&\\
RCAA~\cite{chang_rcaa_2018} & global \& context &RL reward& & 81.3 & & &&\\
I-NET ~\cite{he_end--end_2019} & global &OLP, HEP& 79.5 & 81.5 & & &&\\
MGTS~\cite{chen_person_2018} &global \& mask&OIM& 83.0 & 83.7 & 32.6 & 72.1 &&\\
KD-OIM~\cite{munjal_knowledge_2019} & global &OIM& 83.8 & 84.2 &&&&\\
CGPS~\cite{yan_learning_2019} & global \& context &OIM& 84.1 & 86.5 & 33.4& 73.6 &&\\
PFFN~\cite{hong_scale_2019} &global \& multi scale&Triplet loss& 84.5 & 89.8 & 34.3 & 73.9 &&\\
SMG~\cite{zheng_segmentation_2020} & global \& mask&Binary Cross Entropy& 86.3 & 86.5 & & &&\\
FPSP~\cite{li_fast_2019} &global&Cross entropy& 86.99 & 89.87 & 44.45 & 70.58 &&\\
CLSA~\cite{lan_person_2018} & global \& multi-scale &Cross entropy& 87.2 & 88.5 & 38.7 & 65.0 &&\\
APNet~\cite{zhong_robust_2020} & local &OIM& 88.9 & 89.3 & 41.9 & 81.4 & 18.8 & 55.7\\
DHFF ~\cite{lu_dhff_2019} & global \& multi-scale &Multi-Metric loss& 90.2 & 91.7 & 41.1 & 70.1 &&\\
BINet ~\cite{dong_bi-directional_2020} & global \& local &OIM& 90.8 & 91.6 & 47.2 & 83.4 &&\\
NAE+ ~\cite{chen_norm-aware_2020} & global &OIM& 92.1 & 92.9 & 44.0 & 81.1 &&\\
DKD~\cite{zhang_diverse_2020} &global \& local&& 93.6 & 94.72 & {\bf 54.16} & {\bf 87.89} &&\\
\hline
\multicolumn{9}{c}{\textbf{\textit{Identity-driven detection}}}\\
\hline
NPSM ~\cite{liu_neural_2017} & global &Softmax& 77.9 & 81.2 & 24.2 & 53.1 &&\\
QEEPS ~\cite{munjal_query-guided_2019} & global &OIM& 84.4 & 84.4 & 37.1 & 76.7 &&\\
KD-QEEPS~\cite{munjal_knowledge_2019} & global &OIM& 85.0 & 85.5 &&&&\\
IGPN + PCB~\cite{dong_instance_2020} & global && 90.3 & 91.4 & 47.2 & 87.0 &&\\
RDLR~\cite{han_re-id_2019} &global&Proxy Triplet Loss& 93.0 & 94.2 & 42.9 & 70.2 &&\\
TCTS~\cite{wang_tcts_2020} &global&IDGQ loss& {\bf 93.9} & {\bf 95.1} & 46.8 & 87.5 &&\\
\noalign{
\global\dimen1\arrayrulewidth
\global\arrayrulewidth1pt
}\hline
\noalign{
\global\arrayrulewidth\dimen1 
}
\end{tabular}
}
\caption{Performance of image-based person search methods on CUHK-SYSU, PRW and LSPM datasets.}
\label{table:byImage}
\end{center}
\vspace{-4mm}
\end{table*}

\begin{table*} [t]
\begin{center}
\resizebox{0.8\textwidth}{!}{
\begin{tabular}{l|c|c|c c c}
\multicolumn{1}{c}{} & \multicolumn{1}{c}{} &\multicolumn{1}{c}{} &\multicolumn{3}{c}{{\bf CUHK-PEDES}} \\
\noalign{
\global\dimen1\arrayrulewidth
\global\arrayrulewidth1pt
}\hline
\noalign{
\global\arrayrulewidth\dimen1 
}
{\bf Method} & {\bf Feature} &{\bf Loss}& R@1 & R@5 & R@10\\
\hline
GNA-RNN~\cite{li_person_2017} & global &Cross entropy& 19.05 & & 53.63 \\
CMCE~\cite{li_identity-aware_2017} & global &CMCE loss& 25.94 && 60.48 \\
PWM+ATH ~\cite{chen_improving_2018-1} & global &Cross entropy& 27.14 & 49.45 & 61.02 \\
Dual-Path ~\cite{zheng_dual-path_2020} & global &Ranking loss, Instance loss& 44.4 & 66.26 & 75.07 \\
CMPM+CMPC ~\cite{zhang_deep_2018} & global &CMPM, CMPC& 49.37 & & 79.27 \\
LPS+MCCL~\cite{liu_deep_2019} & global &MCCL& 50.58 & & 79.06 \\
A-GANet~\cite{liu_deep_2019} & global &Binary Cross Entropy& 53.14 & 74.03 & 81.95 \\
PMA~\cite{jing_pose-guided_2020} & global \& pose && 53.81 & 73.54 & 81.23 \\
TIMAM~\cite{sarafianos_adversarial_2019} & global &Cross Entropy, GAN Loss& 55.41 & {\bf 77.56} & {\bf 84.78} \\
ViTAA~\cite{wang_vitaa_2020} & global \& attribute &Alignment loss& {\bf 55.97} & 75.84 & 83.52 \\
\noalign{
\global\dimen1\arrayrulewidth
\global\arrayrulewidth1pt
}\hline
\noalign{
\global\arrayrulewidth\dimen1 
}
\end{tabular}
}
\caption{Performance of text-based person search methods on the CUHK-PEDES dataset. 
}
\label{table:byText}
\end{center}
\vspace{-4mm}
\end{table*}
\section{Datasets and Evaluation}
\begin{table} [t]
\begin{center}
\resizebox{\columnwidth}{!}{
\begin{tabular}{l|ccc|c}
\multicolumn{1}{c}{} & \multicolumn{3}{c}{{\bf Image-Based}} & \multicolumn{1}{c}{{\bf Text-Based}} \\
\noalign{
\global\dimen1\arrayrulewidth
\global\arrayrulewidth1pt
}\hline
\noalign{
\global\arrayrulewidth\dimen1 
}
{\bf Dataset} & CUHK-SYSU & PRW & LSPS & CUHK-PEDES\\
\hline
\#frames & 18184 & 11816 & 51836 & 40206 \\
\#identities & 8432 & 932 & 4067 & 13003\\
\#anno boxes & 96143 & 34304 & 60433 &  \\
\#parts & 6\% & 0\% &60\%\\
\#cameras & & 6 & 17 & \\
\#description &&&&80440\\
\#detector & hand & hand & Faster R-CNN & \\

\noalign{
\global\dimen1\arrayrulewidth
\global\arrayrulewidth1pt
}\hline
\noalign{
\global\arrayrulewidth\dimen1 
}
\end{tabular}
}
\caption{Person search datasets statistics. 
}
\label{table:datasets}
\end{center}
\vspace{-4mm}
\end{table}
\subsection{Datasets}
{\bf \noindent CUHK-SYSU}~\cite{xiao_joint_2017} dataset is an image-based person search dataset, which contains 18184 images, 8432 person identities, and 99809 annotated bounding boxes. The training set contains 11206 images and 5532 query identities. The test set contains 6978 images and 2900 query identities. The training and test sets have no overlap on images or query person. 

{\bf \noindent PRW}~\cite{zheng_person_2017} dataset has a total of 11816 frames which are manually annotated with 43110 person bounding boxes. 34304 people have identifications ranging from 1 to 932, and the rest are assigned identities of -2. The PRW training set has 5704 images and 482 identities, and the test set has 6112 pictures and 450 identities. 

{\bf \noindent LSPS}~\cite{zhong_robust_2020} dataset is a new image-based person search dataset, in which a total number of 51,836 pictures are collected. 60,433 bounding boxes and 4,067 identities are annotated. LSPS has a substantially larger number of incomplete query bounding boxes of 60\% compare to 6\% in CUHK-SYSU and 0\% in PRW.

{\bf \noindent CUHK-PEDES} dataset ~\cite{li_person_2017} is currently the only dataset for text-based person search.  The images are collected from five person re-id datasets and added the corresponding language annotations. It contains 40206 images of 13003 identities and 80440 textual descriptions. Each picture has 2 textual descriptions. The dataset is divided into three parts, 11003 training individuals with 34054 images and 68126 captions, 1000 validation persons with 3078 images and 6158 sentences, and 1000 test identities with 3074 pictures 6156 captions. 

CUHK-SYSU and PRW are de facto datasets for image-based person search. LSPS is new to the community and contains many partial body bounding boxes, making it a specialised dataset to evaluate methods exploiting local discriminative features. CUHK-PEDES is the only text-based person search dataset, and new datasets may further advance research in this area. Dataset statistics are summarised in Table \ref{table:datasets}.

\subsection{Evaluation Metrics}
Cumulative matching characteristics (CMC top-K) and mean averaged precision (mAP) are the primary evaluation metrics for person search. In CMC, the top-K predicted bounding boxes are ranked according to the intersection-over-union (IoU) overlap with the ground-truths equal to or greater than 0.5. The mAP is a popular evaluation metric in object detection, in which an averaged precision (AP) is calculated for each query person, and then the final mAP is calculated as an average of all APs.

\subsection{Performance Analysis}
In this section, we summarise and analyse the evaluation results considering the three significant challenges in person search discussed earlier. We aim to present the influencing factors that contribute to the person search performance. We don't discuss CNN backbones as modern CNN backbones such as ResNet50 and VGG are similar in performance and are mostly interchangeable in different methods. 

We summarise the evaluation results of image-based person search methods in Table \ref{table:byImage}.
We annotate feature types and loss functions used for metric learning along with the methods. Image-based person search faces the steep detection-identification inconsistency challenge. Therefore, we divide image-based person search methods into identity-driven detection and non-identity-driven detection methods to analyse the identity-driven detection solution's effectiveness. 

Methods specifically addressing the detection and identification inconsistency challenge, such as IGPN, RDLR and TCTS, outperform methods addressing the detection and identification separately. Methods exploiting fine-grained discriminative features without considering the detection-identification inconsistency challenge don't have a clear edge over methods using global features. Our interpretation is that the query identity presents in the gallery images. Therefore, the detected person needs to be consistent with the query identity for better query-person matching. For example, if the detected person features are free from noises, the query should be free of noises. Loss functions play critical roles in guiding feature representation learning, such as using a center loss on top of the OIM loss to bring the same identities closer and separate different identities. Knowledge distillation is a notably effective strategy in training the detection and identification models. KD-OIM, KD-QEEPS and DKD beat the corresponding baseline methods without knowledge distillation.

The performance of the text-based person search methods on CUHK-PEDES is summarised in Table \ref{table:byText}. We include feature types and loss functions along with the methods. Text-based person search is essentially a text-image matching problem, and fine-grained discriminative features play a critical role in cross-modal matching. 
Recent methods exploiting fine-grained discriminative features with novel loss functions outperform methods using global features and vanilla Cross-Entropy loss. Specifically, ViTAA~\cite{wang_vitaa_2020} exploiting local discriminative features via attribute-feature alignment achieves the best search results. 
\section{Discussion and Future Directions}
In this survey, we review the recent person search advances covering both image-based and text-based person search. There have been remarkable achievements in the past few years, it remains an open question on addressing the three significant person search challenges, namely the discriminative features, the query-person gap and the detection-identification inconsistency. Next, we discuss a few future research directions. 

{\bf \noindent Multi-modal person search.}
Exiting works focus on search by either image or text. None of them attempted a multi-modal search approach, in which query image and query text complement each other. Multi-modal person search is handy when a partial person image is available such as a passport-sized image. At the same time, the free text provides the rest of the body appearance. Specifically, the CUHK-PEDES dataset can be extended with annotated bounding boxes. Thus CUHK-PEDES has both annotated bounding boxes and textual descriptions, making it a suitable candidate dataset for multi-modal person search. 

{\bf \noindent Attribute-based person search.}
It is a big challenge for a machine to learn complex sentence syntax. Attribute-based person search method AIHM~\cite{dong_person_2019} outperforms the text-based method GNA-RNN~\cite{li_person_2017} evaluated on cropped person images with attribute annotations. Therefore, it's worthwhile to collect attribute annotated scene images and further advance attribute-based person search. The state-of-the-art text-based person search method ViTAA~\cite{wang_vitaa_2020} decomposes textual description to attributes to learn fine-grained discriminative features. Attribute annotations may ease this process and subsequently improve text-based person search performance.

{\bf \noindent Zero-shot person search.}
Text-based person search is essentially a zero-shot learning problem, in which the query person is unseen in training. ~\cite{dong_person_2019} formulates the attribute-based person search as a Zero-Shot Learning (ZSL) problem. In zero-shot learning, zero training image is available at training time, and only semantic representations such as textual descriptions are available to infer unseen classes. Text-based person search can leverage the knowledge of zero-shot learning, such as using adversarially generated person features to augment training data.

\section{Conclusion}
In this survey, we provide a systematic review of the recent works on person search. For the first time, we surveyed papers on text-based person search which is less investigated than image-based person search. We briefly discuss highly regarded methods from the perspective of challenges and solutions. We summarise and compare person search methods' performance and provide insights that a person search method needs to address the joint challenge of discriminative features, query-person gap, and detection-identification inconsistency. We finally discuss some future research directions which may be of interest to incumbent and new researchers in the field.

\bibliographystyle{named}
\bibliography{ijcai21}

\end{document}